\documentclass[letterpaper, 10 pt, journal, twoside]{ieeetran}
\usepackage{xcolor}
\usepackage{soul}
\usepackage{multirow}
\usepackage{graphicx}
\usepackage{booktabs}
\usepackage{dirtytalk}
\usepackage{bm}
\usepackage{hyperref}
\hypersetup{
    colorlinks=true,
    linkcolor=red,
    filecolor=magenta,      
    urlcolor=blue,
    pdftitle={Overleaf Example},
    pdfpagemode=FullScreen,
    }
\usepackage{amsmath}
\usepackage{amssymb}
\usepackage{bbold}
\usepackage{mathtools}
\usepackage{subcaption}
\usepackage{caption}
\usepackage{perl_acronyms}
\let\labelindent\relax
\usepackage[inline]{enumitem}
\usepackage{siunitx}
\usepackage[
    style=ieee,
    doi=false,
    isbn=false,
    url=false,
    eprint=false,
    backend=bibtex,
    natbib=true
    ]{biblatex}
\newcommand{\hlcgreen}[1]{{#1}}
\newcommand{\hlcblue}[1]{{#1}}

\IEEEoverridecommandlockouts                        

\title{Learning Cross-Scale Visual Representations for Real-Time Image Geo-Localization}

\author{Tianyi Zhang$^{1}$, Matthew Johnson-Roberson$^{1}$
\thanks{Manuscript received: September, 9, 2021; Revised January, 3, 2022; Accepted January, 31, 2022.}
\thanks{This paper was recommended for publication by Pauline Pounds upon evaluation of the Associate Editor and Reviewers' comments.} 
\thanks{$^{1}$T. Zhang and  M. Johnson-Roberson are with the Robotics Institute and the Department of Naval Architecture and Marine Engineering,
        University of Michigan,
        Ann Arbor, MI 48109, USA
        {\tt\footnotesize \{tyz, mattjr\}@umich.edu}}%
\thanks{Digital Object Identifier (DOI): see top of this page.}
}

\begin{document}

\maketitle

\begin{abstract}
\urlstyle{rm}
\renewcommand{\thefootnote}{\roman{footnote}}
Robot localization remains a challenging task in GPS denied environments. State estimation approaches based on local sensors, e.g. cameras or IMUs, are drifting-prone for long-range missions as error accumulates. In this study, we aim to address this problem by localizing image observations in a 2D multi-modal geospatial map. We introduce the cross-scale\footnote{\label{footnotescale}Elsewhere in the literature \textit{scale} usually refers to the size of the dataset. However throughout this paper, we use \textit{scale} to refer to the geospatial scale (\textit{pixel/m}) of maps or image observations. \textit{Cross-scale} means that that the dataset contains maps of small scale and image observations of large scale.}\label{footnotescale} dataset and a methodology to produce additional data from cross-modality sources. We propose a framework that learns cross-scale visual representations without supervision. Experiments are conducted on data from two different domains, underwater and aerial. In contrast to existing studies in cross-view image geo-localization, our approach a) performs better on smaller-scale multi-modal maps; b) is more computationally efficient for real-time applications; c) can serve directly in concert with state estimation pipelines. Our code and data are released at \url{https://github.com/tyz1030/CroScaleRep.git}
\end{abstract}

\section{INTRODUCTION}
\label{sec:introduction}
\par \IEEEPARstart{G}{eo-localization} plays a key role in autonomous and robotic systems exploring a priori unknown environments in the wild. 
To achieve better localization accuracy, a wide range of sensors have been used on today's field robots. According to the reference frame, sensors can be generally categorized as local or global. Local sensors, e.g. cameras and \acp{IMU}, observe the environment in a local coordinate frame. Global sensors, e.g. \acp{GPS}, barometers, and magnetometers, provide global measurements in fixed global frames.
While local sensors give high-precision local measurements, global sensors are noisier but do not suffer from the same drift effects when localizing the vehicle.
The algorithmic combination of both kinds of sensors achieves locally accurate and globally drift-free performance on long-range tasks~\cite{Qin2019AGO}.


\par However, there are many scenarios where global information is not available or only partially available. The scenarios can be underwater, underground, or other \ac{GPS} denied environments. Taking underwater as an example, neither \ac{GPS} nor land-based station towers can be accessed since electromagnetic waves are heavily attenuated. Acoustic localization gets downgraded by variation in salinity or temperature in the water body. Magnetometer and depth sensors are reliable global sensors underwater, however, they only provide measurements up to 4 \acp{DOF} in total. The global measurements of the most important 2 \acp{DOF} on the horizontal plane are missing anyway.

\par The absence of global sensing has raised a global challenge for geo-localization: the incremental state estimation based on local-only sensor systems, i.e. dead reckoning, is prone to drifts which accumulate with time. Hence, in long-range missions, we need to find an approach to control the growing localization error online.

\begin{figure}[t]%
\centering
    \begin{subfigure}{.48\textwidth}
    \includegraphics[width=\linewidth]{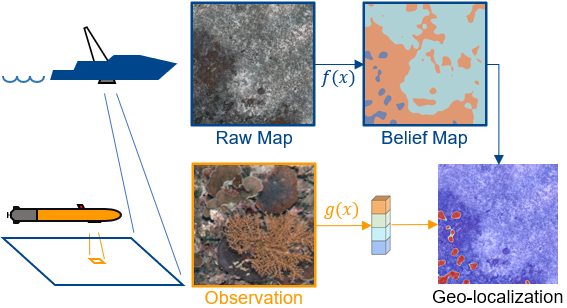}
    \end{subfigure}%
    \caption{Cross-scale\hlcgreen{\textsuperscript{\ref{footnotescale}}} geo-localization: Small-scale raw map and large-scale image observation are encoded into probabilistic representations. Location of the image observation can then be inferred in the map.}%
    \label{intro:concept}
\end{figure}

\par In this study, we conceive a real-time geo-localization system (see Fig.~\ref{intro:concept}) for robot platforms equipped with RGB cameras. In this system, a 2D geospatial map is encoded into a belief map. Observations from the camera are encoded into feature representations and its 2D location can be inferred in the belief map. The key problem we address in this study is how to encode 2D geospatial maps of various modalities and image observations into consistent representations. The contributions of this study are as follows:
\begin{itemize}
    \item We formulate the cross-scale\hlcgreen{\textsuperscript{\ref{footnotescale}}} geo-localization task, which can help robots with large-scale image observations navigate in small-scale maps.
    \item We propose a workflow to sample cross-scale visual data from cross-modality resources. We build and release two datasets from different domains.
    \item We propose to use a pixels-to-pixels map encoder to efficiently encode map patches with small scale.
    \item We propose a framework which trains a map encoder and an image observation encoder jointly without supervision. We use Bhattacharyya coefficient~\cite{bhattacharyya1946measure} as the similarity metric between the probabilistic outputs of both encoders. We modified NT-Xent~\cite{chen2020simple} as loss function for our case.
    \item We propose to use Dirichlet distribution to model the probability of encoded observation in an encoded map, which can be potentially leveraged in downstream inference applications.
\end{itemize}

\section{RELATED WORK}
\label{sec:relatedwork}
\par \subsection{\Ac{TAN} with particle filtering} Navigating an \ac{AUV} with a terrain elevation map has been studied for decades. Early work was conducted on an \ac{AUV} equipped with single-point sonar and water depth sensor~\cite{uwpf2003}. Since acquiring information from a map is a highly non-linear operation, a particle filter method is applied to \ac{TAN} for state estimation. Subsequent work improved the \ac{TAN} method by plugging in different kinds of range sensors~\cite{tan2008profile, tan2019multibeam}, upgrading the particle filter into different variants~\cite{tan2008pmf, tan2017pfcompare, tan2019mpmf}, developing efficient mapping features~\cite{tan2006floodslam, tan2009effslam, tan2010slam}, and realizing cooperative \ac{TAN} with multiple vehicles~\cite{tan2016cooperative, tan2020colla, tan2021communication}. However, the success of a \ac{TAN} system requires sufficient excitation from terrain elevation, which is not always guaranteed in many scenarios.
\par In our work, we break the limitation of using range sensors and a terrain elevation map. Instead, we explore the possibility of using RGB cameras as sensors, and maps of different modalities which provide richer geospatial information.

\par \subsection{Cross-view image geo-localization} Cross-view image geo-localization refers to determining the geolocation of a query image with an overhead satellite image. This problem was first formulated as an information retrieval problem in ground-and-overhead image databases and was attempted based on the extraction and matching of hand-crafted features~\cite{cros2013lin}.
Workman et al.~\cite{cros2015workman} and 
Vo and Hays~\cite{vo2016localizing} approached this problem with \ac{CNN} backbones for different views and evaluated the performance of different training strategies and embedding architectures.
Hu et al.~\cite{cros2018cvmnet} proposed CVM-Net which embedded the NetVLAD layer~\cite{cros2018netvlad} on top of a \ac{CNN} to extract descriptors invariant to viewpoint changes. Other extensive studies~\cite{cros2019airgan, cros2019safa, cros2020WhereAI} developed domain transfer methods to bridge the gap between different viewpoints, but only apply to the cases that query image is panoramic. 
\par Methods developed in this field of study are all based on the end-to-end framework of information retrieval and ranking. This makes it difficult to integrate such systems in \ac{TAN} or other general \ac{SLAM} workflow, which are mostly based on filtering and optimization. Moreover, the exhaustive sliding window search lacks the efficiency to deploy on the mobile platforms need real-time localization in a dynamically updating map.
\par In our study, we move away from the information retrieval framework and develop solutions more efficient and compatible for real-time geo-localization.

\par \subsection{\Ac{RS} image classification} \par Studies in \ac{RS} image classification have inspired us with the feature association problem. Cao et al.~\cite{rs2018cao} introduced a land use classification network with both aerial and street view images integrated. Street view features are interpolated by geo-coordinates and concatenated with aerial features. Hong et al.~\cite{hong2020more} proposed feature fusion and network training strategies for the multi-modal and cross-modal \ac{RS} image classification. Above mentioned approaches both need supervision from ground truth labels to train.

\par \subsection{Contrastive learning} \par Recent progress in contrastive learing has shown how \ac{CNN}s learn visual representations without supervision~\cite{chen2020simple, he2020momentum}. Further, Pielawski el al.~\cite{pielawski2020comir} proposed \ac{CoMIR} which addressed the multimodal image registration problem with contrastive learning. However, \ac{CoMIR} works with different modalities of exactly the same scale, which means that it is not directly applicable to our localization problem which has a large scale ratio between image observations and the map.


\section{PROBLEM FORMULATION}
\label{sec:problem}
\par We aim to address the geo-localization problem with RGB image observations and multi-modal maps referenced in 2D geo-coordinates. RGB camera provides rich visual information and is one of the most affordable and widely-equipped perceptual sensors on mobile robots.
However, high-resolution RGB satellite images which serve as maps in existing studies have limited coverage on the earth. We seek to exploit other lower-resolution modalities with smaller scale to serve as a map.
In contrast to existing image geo-localization approaches, we focus on following goals:
\renewcommand{\theenumi}{\roman{enumi}}
\begin{enumerate}     
    \item Localization in smaller-scale maps (typically, scale of an image observation has a magnitude of $\times 10^1$ or larger than a map);
    \item Localization in maps of different modalities, which massively extend the data that can serve as maps;
    \item Efficient computation for real-time data processing;
    \item Compact map description for potentially efficient transmission over network;
    \item Compatibility as a plug-in module (instead of an end-to-end standalone program) in state estimation pipelines. 
\end{enumerate}

\section{METHODOLOGY}
\label{sec:methodology}
\subsection{Cross-Scale Dataset}
\par The basic data unit of our proposed cross-scale dataset is a data tuple, which consists of a 2D map patch $M$, $n$ image observations $O=\{o_j\}_{j=1}^n$ and $n$ pixel coordinates $P=\{(u_j, v_j)\}_{j=1}^n$ indicating where elements of $O$ are located in $M$. The whole dataset consists of a certain number of such data tuples sampled from one or multiple areas of interest.
\par
\begin{figure}[t]%
\centering
    \begin{subfigure}{.44\textwidth}
    \includegraphics[width=\textwidth]{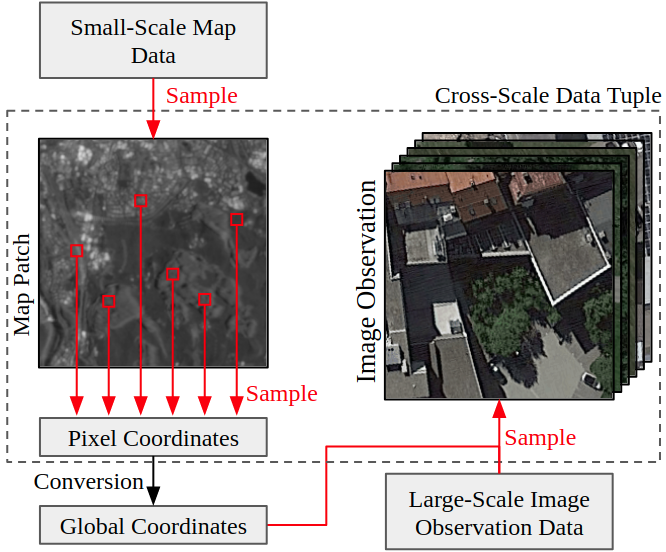}
    \end{subfigure}%
    \caption{Workflow of sampling the cross-scale dataset from data source of map and image observations. As the basic unit of our proposed dataset, a data tuple consists of a map patch, certain number of image observations and the pixel coordinates in the map patch \hlcblue{where the center of image observations are located}.}%
    \label{meth:samp_flow}
\end{figure}

 Fig.~\ref{meth:samp_flow} shows the workflow of sampling a data tuple $(M,O,P)$. First, we sample a map patch $M$ with random coordinate and random rotation from the data source of small-scale map.
Then we randomly sample $P$ from map patch as the \hlcblue{centers} where $O$ will be sampled.
$P$ will be converted into global coordinates for sampling $O$ from the data source of large-scale images.

\begin{figure}[t]%
\centering
    \begin{subfigure}{0.80\linewidth}
    \includegraphics[width=\linewidth]{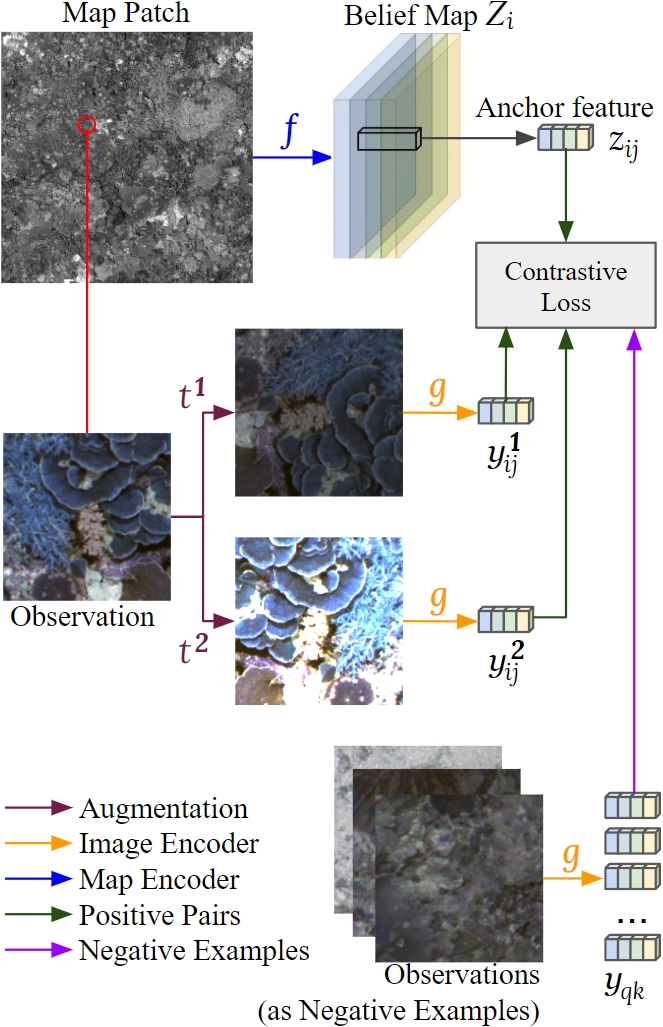}
    \end{subfigure}%
    \caption{The overview of our proposed framework. A map patch is encoded into a belief map. An image observation is augmented into two views then encoded into 1D representation. Anchor features are extracted from belief map by pixel coordinates. Image's representations and anchor features with corresponding pixel coordinates will serve as positive pairs and the rest of the images will be the negative examples.}%
    \label{meth:network}
\end{figure}

\subsection{Network}
\par The proposed network consists of a map encoder $f$ and a observation encoder $g$ to extract features from map patches and image observations (see Fig.~\ref{meth:network}). $f$ encodes a map patch into a belief map, which has the same height and width with the input. The belief map has $C$ channels, which corresponds to the $C$ categories of terrain representations. $g$ encodes the image into a 1D representation of size $C$.
Both encoders are expected to learn representations consistent for the same location while distinguishable for different types of terrain.
\par Contrastive learning is applied for jointly training both encoders.  For clarity, in this paper we use $i$ to index data tuples in a minibatch, and $j$ to index image observations and their corresponding pixel coordinates in one data tuple. Within a mini-batch $B = \{(M_i,O_i,P_i)\}_{i=1}^b$ of size $b$, each map patch $M_i$ is encoded into a belief map $Z_i = f(M_i)$.
No matter how $f$ is realized, $softmax$ function as last layer will convert the output score of each pixel into a discrete probabilistic distribution.
We expect that potential downstream applications do inference based on this property.
The anchor feature $z_{ij}$ is then extracted from ${Z_i}$ by coordinate $P_i$, denoted by $z_{ij}={Z_i}[u_{ij}, v_{ij}]$.
Each image observation $o_{ij}$ is first randomly augmented into 2 views, $t^\mathbf{1}(o_{ij})$ and $t^\mathbf{2}(o_{ij})$, where $t^\mathbf{1}$ and $t^\mathbf{2}$ are different augmentation operators randomly sampled from same augmentation family. 
While various kinds of image augmentation has been recommended by~\cite{chen2020simple}, it's up to the task which augmentation will be applied in training. For example, color distortion can be necessary for underwater applications. Let $y^\mathbf{1}_{ij} = g(t^\mathbf{1}(o_{ij}))$ and $y^\mathbf{2}_{ij} = g(t^\mathbf{2}(o_{ij}))$ be the
representations encoded from both augmented views. Similar to $f$, $softmax$ operation is applied as the last layer of $g$. Both $y^\mathbf{1}_{ij}$ and $y^\mathbf{2}_{ij}$ are considered as positive examples of the anchor feature $z_{ij}$. We treat augmented $2(b\cdot n-1)$ views of all the rest image observations in the mini-batch as negative examples.

\subsection{Similarity and Contrastive Loss}
We adapt NT-Xent (the normalized
temperature-scaled cross entropy loss)~\cite{chen2020simple} to our case where one anchor feature has two positive examples. The loss function $\mathcal{L}_{ij}$ for an anchor feature $z_{ij}$ is defined as:

\begin{equation}
    \begin{aligned}
    \mathcal{L}_{ij} =-\sum\limits_{\lambda = \mathbf{1}}^{\mathbf{2}} log{\frac{e^{s(z_{ij}, y^{\lambda}_{ij})/\tau}}{\sum\limits_{q=1}^b \sum\limits_{k=1}^n\sum\limits_{\mu = \mathbf{1}}^{\mathbf{2}} e^{s(z_{ij}, y^\mu_{qk})/\tau}}}
    \end{aligned}
\end{equation}
where $\tau$ is the temperature parameter~\cite{chen2020simple}. $s(z, y)$ is the similarity function between $z$ and $y$. Since the output after $softmax$ layer are interpreted as discrete probabilistic distributions, we use Bhattacharyya coefficient~\cite{bhattacharyya1946measure} as the similiarity function
:
\begin{equation}
    s(z, y) = \sqrt{z}^\top\sqrt{y}
\end{equation}
where $\sqrt{\ }$ performs element-wise operation. Though cosine similarity is widely used in well-proved contrastive learning frameworks~\cite{chen2020simple, he2020momentum}, we do not use it in this work. Cosine similarity normalizes the encoder outputs before $softmax$, removing the ability to constrain the magnitude of the network activation. Such property will lead to inconsistent outputs between two encoders after $softmax$.


\subsection{Inference}
\par At location $[u, v]$ (pixel coordinate in belief map), we want to find the distribution of observing representation ${y}$ given ${Z}[u, v]$ from belief map. We model the distribution of observed representation as a Dirichlet Distribution:
\begin{equation}
p({y}|{Z}[u, v]) \sim dirichlet(\textbf{1}+\theta {Z}[u, v])
\label{eq:diri}
\end{equation}
where $\textbf{1}=[{1,\dots, 1, 1}]$ has the same size with ${y}$. $\theta$ is a free parameter to tune.
\hlcgreen{Given an observed representation $\hat{y}$, The likelihood $L([u,v], \hat{y})$ is modeled by:}
\begin{equation}
    {{L([u,v], \hat{y})\propto p({y=\hat{y}}|{Z}[u, v])}}
    \label{meth:lklhd}
\end{equation}
\hlcgreen{which can be plugged into general SLAM frameworks. For example, Eq.~\mbox{\ref{meth:lklhd}} can be used to define the \texttt{NonlinearFactor} class as a measurement in the GTSAM framework~\mbox{\cite{dellaert2012factor}}.}

\section{EXPERIMENTS}
\label{sec:experiments}
\subsection{Dataset}
\par We build two datasets for this study, Scott Reef dataset and Kempten dataset. Each dataset contains 1000 data tuples to split for training and validation, and 200 data tuples for testing. For both datasets, we sample the image observation with the resolution of $224\times 224$ and the map patch with $512\times 512$.

\par \textbf{Scott Reef dataset} is built from the Scott Reef 25 dataset (2009)~\cite{acfr2010dataset} provided by \ac{ACFR}. \hlcgreen{Since the raw Scott Reef data are images collected in a ``lawn mowing" pattern without geo-tags, we use the 2D scene reconstruction map from the image sequence to generate the dataset for this study. The camera poses recovered with modern scene reconstruction techniques typically has a guaranteed accuracy and can serve as ground truth for studying localization problems~\mbox{\cite{aqualoc}}. Therefore, in the place of global geo-coordinates, we use the local geo-coordinate of the reconstruction map.} The 2D RGB reconstruction covers an area of approximately $100m\times 100m$ with the resolution $50k\times 50k$. Image observations are sampled from the raw-resolution reconstruction with $\sim500\ pixel/m$ scale. Map patches are sampled from the reconstruction down-scaled by a factor of $\times 8$, which has a scale of $\sim 61\ pixel/m$. \hlcgreen{With this dataset, we will show how our proposed method help robots improve localization over an area with map, which can play an important role on underwater robots surveying a pre-visited area or robot swarm streaming the map to each other.
Potential cross-modality data (e.g. sonar scan or acoustic backscattering with global geo-tags) can be incorporated into our dataset by image registering~\mbox{\cite{pielawski2020comir}} with the 2D reconstruction map.}
\par \textbf{Kempten dataset} is built with the Sentinel-1 SAR~\cite{ee_sentinel} as map and Google Map Satellite~\cite{gmap} as image observations. Sentinel-1 data has a scale of $0.1\ pixel/m$, and 4 channels, i.e. $vv$, $vh$, $hh$ and $hv$. We project the Sentinel-1 data using Pseudo-Mercator (EPSG:3857)~\cite{ee_sentinel}. Google Map Satellite's RGB images have a scale of $\sim 7\ pixels/m$ with zoom level $20$. Data are collected around $longtitude\in [10.22, 10.46]$ and $latitude\in [47.62, 47.77]$. We choose this area because of diverse terrain types and consistent satellite image quality.

\subsection{Implementation}
\par First parameter to be determined is the size $C$ of the output feature, which should be sufficient to describe the types of terrain in the dataset. 
Practically, $C$ will be tuned as a hyperparameter in training. However, the magnitude of $C$ can be predetermined with some human knowledge of terrain types. For example, if we believe the number of terrain types in the area of interest lies between $5$ to $10$, then this will be the range we use to tune the parameter.

\par We realize the map encoder with FCN-ResNet50~\cite{long2015fully}. We define the last \texttt{conv2d} layer of \texttt{FCNHead} to output feature of $C = 5$ channels. We add a \texttt{softmax} layer at the end of the network to convert the score map to the belief map. We choose ResNet18~\cite{he2016resnet} as the image encoder. Similarly, we have the \texttt{FC} layer to output representations of size $C = 5$, and add a \texttt{softmax} layer as the last layer.
\par We use $\tau=1$ for loss function in presenting the results, which is selected empirically.
\par In our image augmentation family, we use the $\mathcal{C}_4$~\cite{pielawski2020comir} (the finite, cyclic, symmetry group of multiples of $90^\circ$ rotations) for learning rotation-invariant representations. We also apply random changes in brightness, contrast, saturation, and hue.
\par The network is trained with SGD optimizer with momentum~\cite{sutskever2013momentum} for 300 epochs. The learning rate starts at $0.002$, and is reduced by a factor of $0.1$ once learning stagnates. We include 8 data tuples in a mini-batch, and each data tuple contains 6 image observations. In other words, a mini-batch contains 8 map patches and 48 image observations, which is the largest that fits in a 12GB GPU.

\subsection{Visualizations}

\begin{figure}[t!]
\centering
    \begin{subfigure}[t]{\linewidth}
      \includegraphics[width=\linewidth]{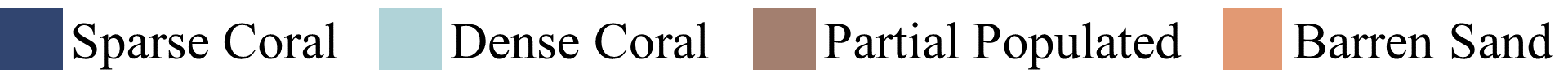}
    \end{subfigure}
    \hfill
    \begin{subfigure}[t]{0.47\linewidth}
      \includegraphics[width=\linewidth]{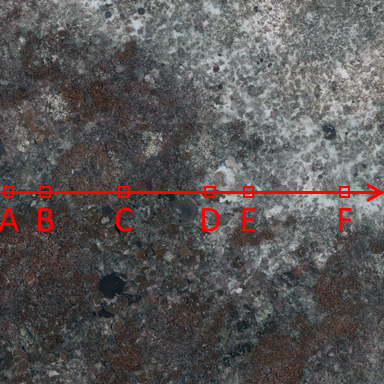}
      \caption{Map patch with RGB channels (scale$\approx$61 pixel/m)}
      \label{exp:fea_scott_raw}
    \end{subfigure}
    \hfill
    \begin{subfigure}[t]{0.47\linewidth}
      \includegraphics[width=\linewidth]{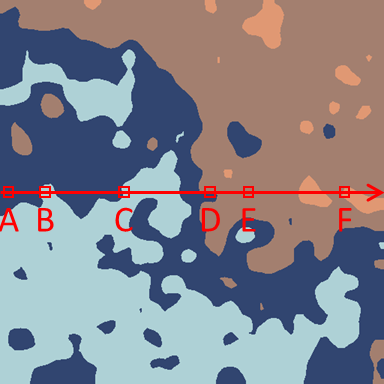}
      \caption{Belief map visualized in segmentation style}
      \label{exp:fea_scott_seg}
    \end{subfigure}

    \begin{subfigure}[t]{\linewidth}
        \centering
        \includegraphics[width=\linewidth]{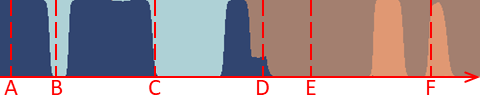}
        \caption{Profile of the belief map along the axis in red}
      \label{exp:fea_scott_profile}
    \end{subfigure}
    \begin{subfigure}[t]{\linewidth}
    \centering
    \includegraphics[width=\linewidth]{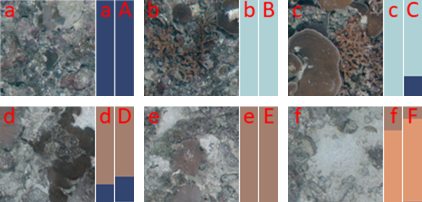}
    \caption{Image observations (scale$\approx$500 pixel/m) and representations, labelled with \textbf{a}-\textbf{f}. Corresponding representations extracted from belief map are labelled with \textbf{A}-\textbf{F}.}
      \label{exp:fea_scott_obs}
    \end{subfigure}
    \caption{Visualized representations for Scott Reef dataset. \hlcgreen{The legend on the top interprets each active channel in learned representations into 4 categories: sparse coral (a/A), dense coral (b/B and c/C), partial populated substrate (d/D and e/E) and barren sand (f/F). Note that both d/D and f/F are combinations of two types of terrains as represented.}}
    \label{exp:feavis_scott}
\end{figure}

\begin{figure}[t!]
\centering
    \begin{subfigure}[t]{\linewidth}
      \includegraphics[width=\linewidth]{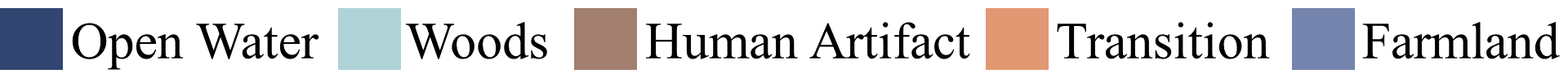}
    \end{subfigure}
    \hfill
    \begin{subfigure}[t]{0.47\linewidth}
      \includegraphics[width=\linewidth]{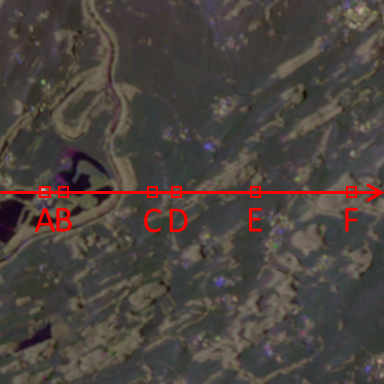}
      \caption{Map patch with hh, hv and vv channels (scale=0.1 pixel/m)}
      \label{exp:fea_kempten_raw}
    \end{subfigure}
    \hfill
    \begin{subfigure}[t]{0.47\linewidth}
      \includegraphics[width=\linewidth]{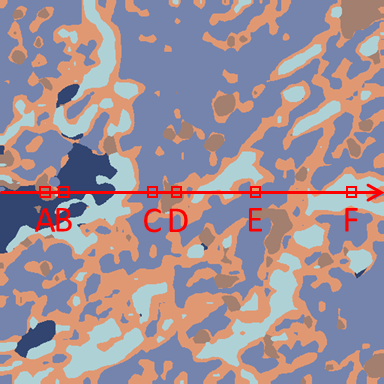}
      \caption{Belief map visualized in segmentation style}
      \label{exp:fea_kempten_seg}
    \end{subfigure}
    \begin{subfigure}[t]{\linewidth}
        \centering
        \includegraphics[width=\linewidth]{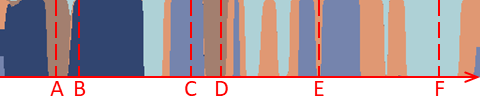}
        \caption{Profile of the belief map along the axis in red}
      \label{exp:fea_kempten_profile}
    \end{subfigure}
    
    \begin{subfigure}[t]{\linewidth}
    \centering
    \includegraphics[width=\linewidth]{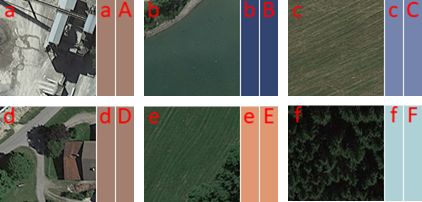}
    \caption{Image observations (scale$\approx$7 pixel/m) and representations, labelled with \textbf{a}-\textbf{f}. Corresponding representations extracted from belief map are labelled with \textbf{A}-\textbf{F}.}
      \label{exp:fea_kempten_obs}
    \end{subfigure}
    \caption{Visualized representations for Kempten dataset. \hlcgreen{The legend on the top interprets each active channel in learned representations into 5 categories: human artifacts (a/A and d/D), open water area (b/B), farmland (c/C), transition area (e/E), and woods (f/F).}}
    \label{exp:feavis_kempten}
\end{figure}

We visualize the representations learned by both backbones to show that they learn consistent representations, see Fig.~\ref{exp:feavis_scott} and Fig.~\ref{exp:feavis_kempten}. Learned representations are colorred channel-wise. Fig.~\ref{exp:fea_scott_raw} shows an RGB map patch from the Scott Reef dataset. Fig.~\ref{exp:fea_scott_seg} presents the belief map in segmentation style by performing $argmax$ operation along the channel dimension.
As shown in 4 different colors, the map patch is encoded into 4 different types of representations.
The segmentation-style map shows a terrain pattern that generally aligns with the raw RGB map.
Fig.~\ref{exp:fea_scott_profile} profiles the belief map along the axis in red.
We can observe changes in 4 types of representations along the axis. \hlcgreen{This profile also reveals a sparse data structure in the belief map, where one channel dominates and the rest have values all close to $0$. Such sparsity guarantees us the potential of compressing the belief map into a compact size.}
We compare the representations encoded from image observation and map in Fig.~\ref{exp:fea_scott_obs}.
The positions where the image observations are located are also indicated in the Fig.~\ref{exp:fea_scott_raw},~\ref{exp:fea_scott_seg} and~\ref{exp:fea_scott_profile}.
From the comparison we see both encoders learn similar representations across scales.
We also find that those representations learned without supervision can be interpreted into description with human oversight\hlcblue{, including sparse/dense coral, partial populated substrate and barren sand (see legend on the top of Fig.~\mbox{\ref{exp:feavis_scott}}).} We do the same visualization for Kempten dataset (Fig.~\ref{exp:feavis_kempten}). In this patch, we see 5 different kinds of terrains represented: \hlcblue{open water, woods, human artifacts, transition area and farmland (see legend on the top of Fig.~\mbox{\ref{exp:feavis_kempten}}).}

\par
\begin{figure*}[t!]
\centering
    \begin{subfigure}[t]{0.15\linewidth}
      \includegraphics[width=\linewidth]{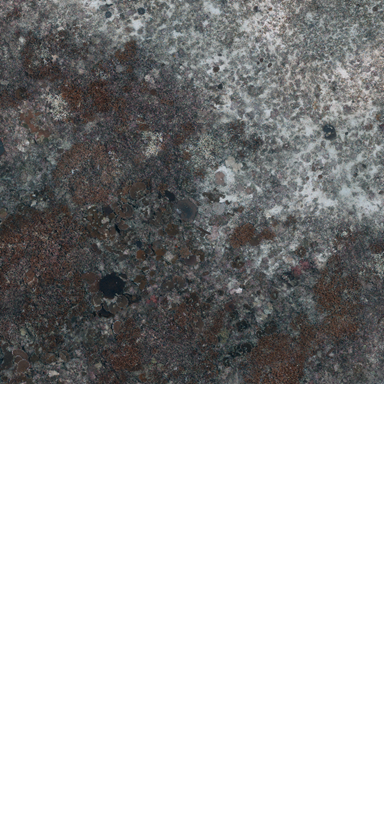}
      \caption{Raw RGB map}
    \end{subfigure}
    \hfill
    \begin{subfigure}[t]{0.15\linewidth}
      \includegraphics[width=\linewidth]{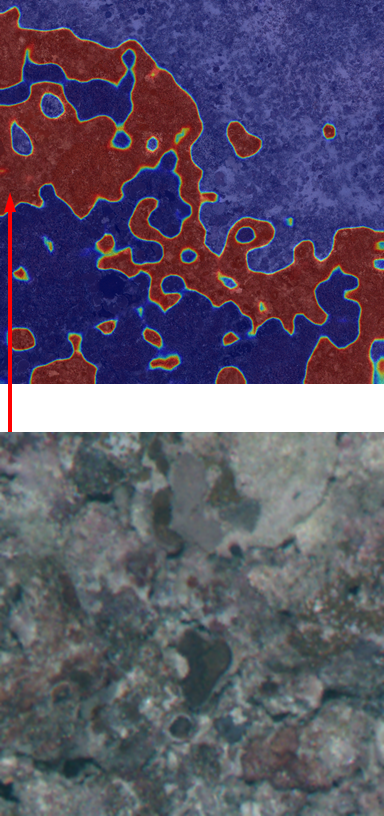}
      \caption{}
    \end{subfigure}
    \hfill
    \begin{subfigure}[t]{0.15\linewidth}
      \includegraphics[width=\linewidth]{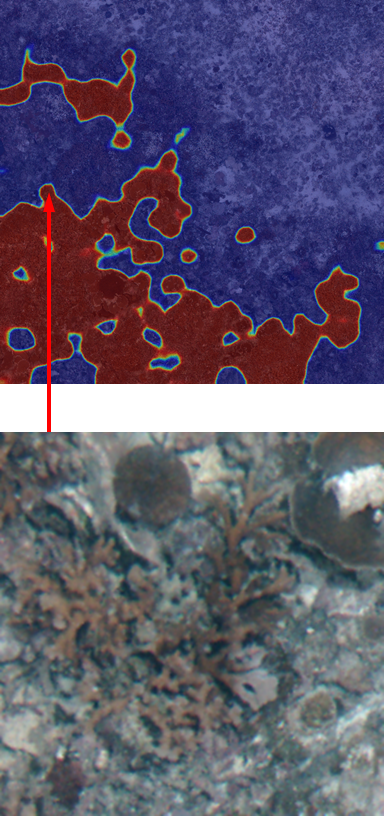}
      \caption{}
    \end{subfigure}
    \hfill
    \begin{subfigure}[t]{0.15\linewidth}
      \includegraphics[width=\linewidth]{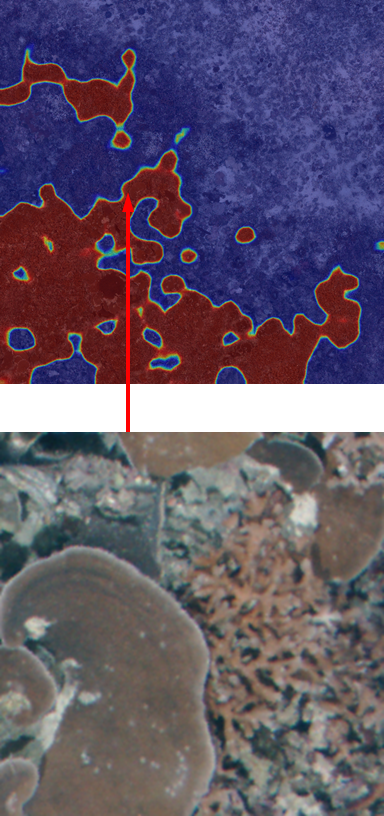}
      \caption{}
    \end{subfigure}
    \hfill
    \begin{subfigure}[t]{0.15\linewidth}
      \includegraphics[width=\linewidth]{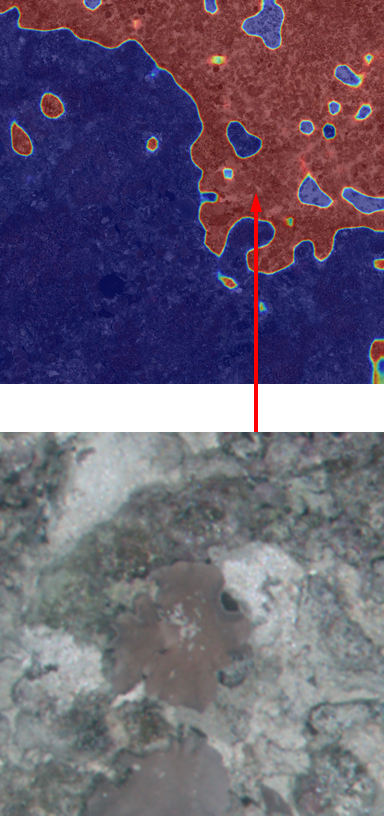}
      \caption{}
    \end{subfigure}    
    \hfill
    \begin{subfigure}[t]{0.15\linewidth}
      \includegraphics[width=\linewidth]{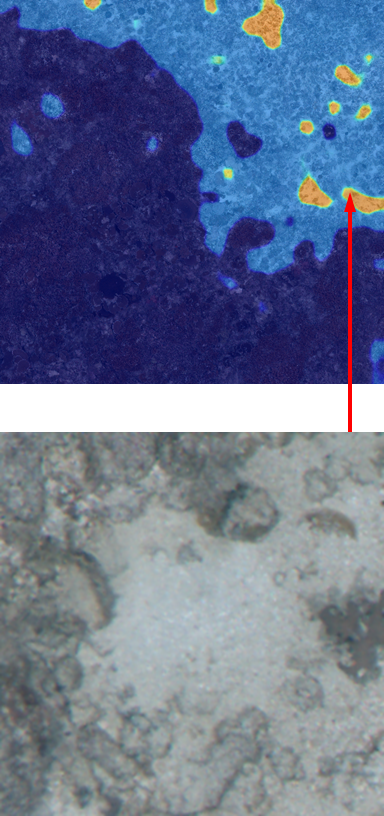}
      \caption{}
    \end{subfigure}
    \caption{Scott Reef Dataset: locations of image observations inferred in the map}
    \label{exp:heat_scott}
\end{figure*}

\begin{figure*}[t!]
\centering
    \begin{subfigure}[t]{0.15\linewidth}
      \includegraphics[width=\linewidth]{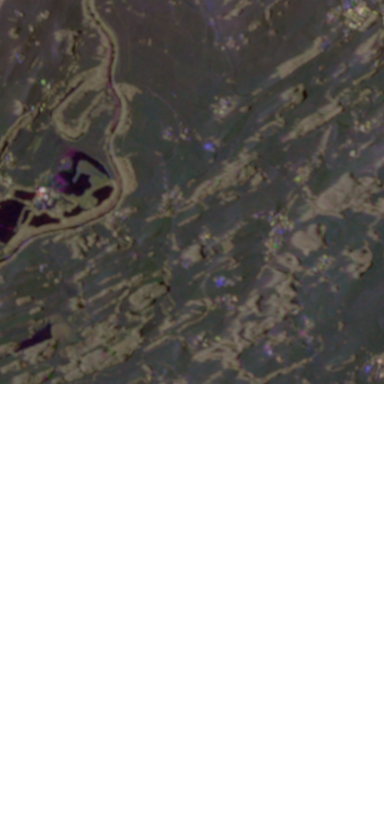}
      \caption{Raw SAR map}
      \label{exp:heat_scott_raw}
    \end{subfigure}
    \hfill
    \begin{subfigure}[t]{0.15\linewidth}
      \includegraphics[width=\linewidth]{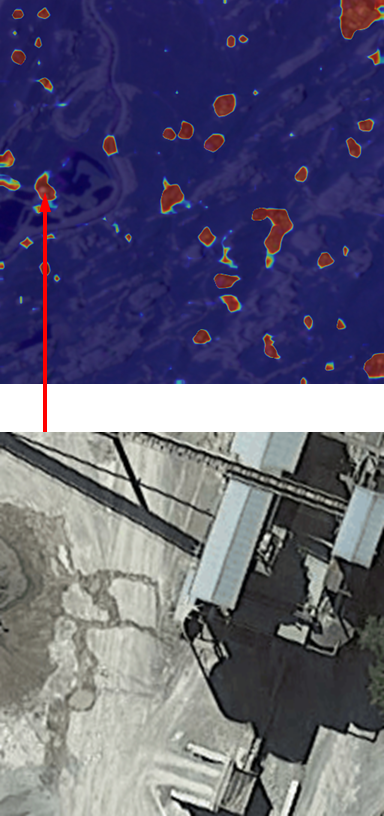}
      \caption{}
    \end{subfigure}
    \hfill
    \begin{subfigure}[t]{0.15\linewidth}
      \includegraphics[width=\linewidth]{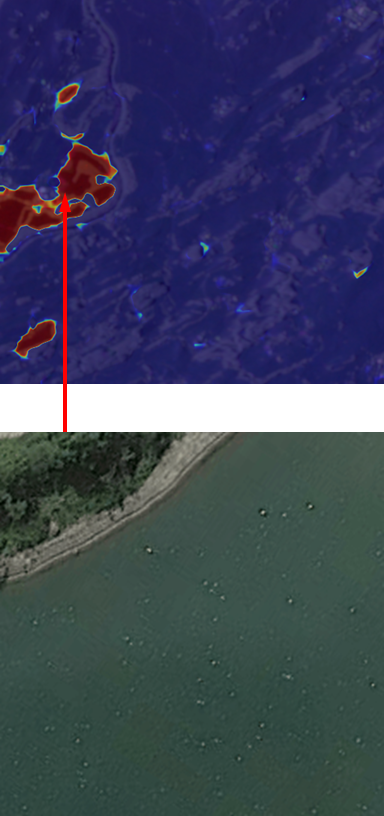}
      \caption{}
    \end{subfigure}
    \hfill
    \begin{subfigure}[t]{0.15\linewidth}
      \includegraphics[width=\linewidth]{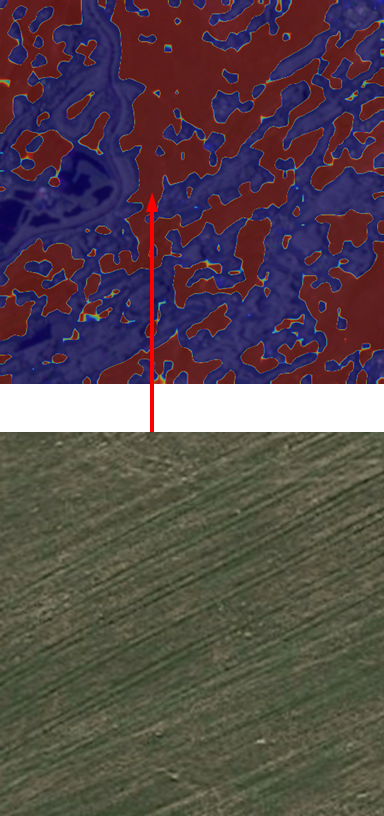}
      \caption{}
    \end{subfigure}
    \hfill
    \begin{subfigure}[t]{0.15\linewidth}
      \includegraphics[width=\linewidth]{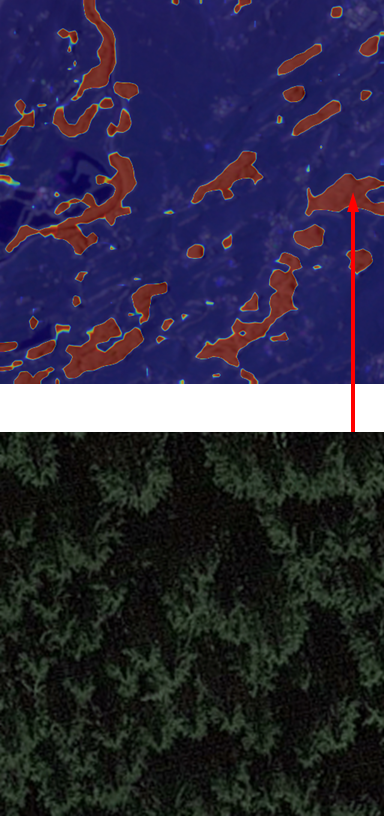}
      \caption{}
    \end{subfigure}
    \hfill
    \begin{subfigure}[t]{0.15\linewidth}
      \includegraphics[width=\linewidth]{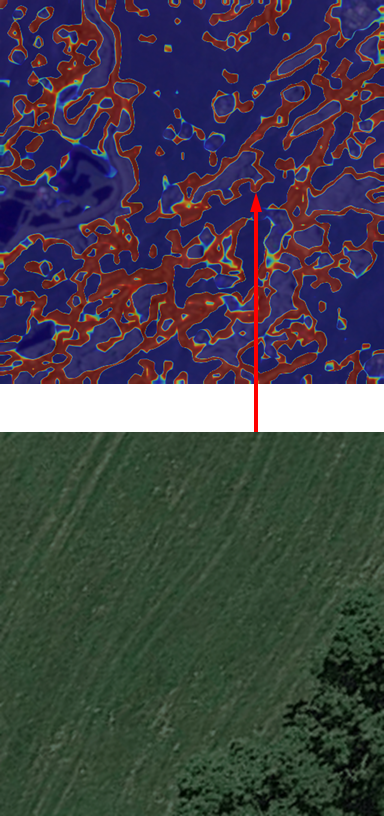}
      \caption{}
    \end{subfigure}
    \caption{Kempten Dataset: locations of image observations inferred in the map}
    \label{exp:heat_kempten}
\end{figure*}

 We evaluate the probability density of image observation with each pixel in the belief map by Eq.~\ref{eq:diri} (with $\theta=5$), and visualize as a heat map (Fig.~\ref{exp:heat_scott} and Fig.~\ref{exp:heat_kempten}). The arrows indicate the ground truth position of image observations in the map. We can see that the images presented all lie in the area with deep red color indicating a high probability density.

\subsection{Numerical Evaluation and Comparisons}
We compare our approach with Triplet Network~\cite{vo2016localizing} and CVM-Net I~\cite{cros2018cvmnet}. To keep the comparison fair, we also experimented with their backbone replaced with ResNet18, which is identical to our image encoder (without the $softmax$ layer).
Since the other approaches all need image pairs from both scales for training and a sliding window search for testing, we sampled small-scale image tiles with resolution $32\times32$ from the raw map. We choose this resolution because it will cover approximately the same (or larger) \ac{FOV} under the scale ratio and the performances of backbone networks are proven on datasets with $32\times 32$ resolution, e.g. CIFAR-10~\cite{Krizhevsky2009cifar10}.

\subsubsection{Recall} We first investigate top-$k\%$ recall rate following~\cite{vo2016localizing, cros2018cvmnet}. A higher recall rate means that the ground truth location is more likely to be included in the area with top-$k\%$ response as inferred. Since we expect our approach to work in conjunction with other localization frameworks as opposed to alone, we evaluate the recall rate on each map patch from the testing set, instead of the whole region of interest. The recall at $1\%$ and $5\%$ are reported in Table~\ref{tab:recall}. Our approach shows second to highest recall performance on Scott Reef dataset and the highest on Kempten dataset. It is worth mention that, the scale ratio between image observation and the map patch is $\times 8$ for Scott Reef, while approximately $\times 70$ for Kempten.
\begin{table}
\centering
\begin{tabular}{c|cc|cc} 
\toprule
\multicolumn{1}{l|}{} & \multicolumn{2}{c|}{Scott Reef}      & \multicolumn{2}{c}{Kempten}          \\
\multicolumn{1}{l|}{} & 1\%  & 5\% & 1\% & 5\%  \\ 
\midrule
Triplet~\cite{vo2016localizing}  & 7.11\%    & 13.92\%   & 18.10\%    & 22.40\%        \\
Triplet  (ResNet18)   & \textbf{63.86\%}  & \textbf{66.98\%}    & 34.91\%  & 37.56\%   \\
CVM-Net~\cite{cros2018cvmnet}    & 29.48\%   & 51.31\%   & 33.61\%   & 38.66\%   \\
CVM-Net (ResNet18)   & 27.27\%  & 48.85\%   & 24.48\%    & 34.42\%
\\Ours   & 51.92\% & 56.35\% & \textbf{35.81\%} & \textbf{40.28\%}  \\
\bottomrule
\end{tabular}
\caption{Comparision by average recall rate}
\label{tab:recall}
\end{table}
The results above imply that as the scale ratio gets larger, it's harder to extract and associate local features across scales. Since our approach observes the map as a whole instead of with a local sliding window, it seems to be the least downgraded approach regarding the recall rate.
\begin{figure*}
    \begin{subfigure}[t]{.48\linewidth}
      \includegraphics[width=\linewidth]{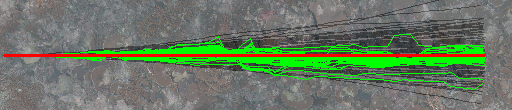}
      \caption{Scott Reef dataset}
    \end{subfigure}
    \hfill
    \begin{subfigure}[t]{.48\linewidth}
      \includegraphics[width=\linewidth]{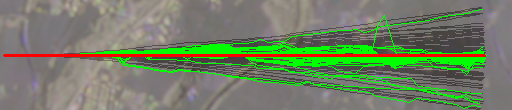}
      \caption{Kempten dataset}
    \end{subfigure}
    \caption{Running particle filter on selected map patches. The groundtruth trajectory, noisy estimations and filtered estimations are plotted in red, grey and green respectively. The trajectory propagates from left to right.}
    \label{exp:pf_img}
\end{figure*}

\subsubsection{Synthetic trajectory}
Comparing to the urban environments studied in~\cite{vo2016localizing, cros2018cvmnet, cros2018netvlad}, features in nature are more repetitive. Particularly in small-scale maps, the network learns similar representations for terrains with the same appearances. Hence we do not report any precision metrics which heavily depend on the unequal coverage of each terrain type.
Instead, we directly evaluate how effective the learned representations are in state estimation with synthetic trajectories.
\hlcblue{We implemented the particle filter based on a \mbox{\ac{TAN}} framework~\mbox{\cite{uwpf2003}} which corrects the state estimation every time an image observation comes in. Instead of simulating the vehicle dynamics, we generate a straight-line 2D trajectory and add Gaussian noise in the incremental estimations. We randomly generate 100 noisy sequences aiming to evaluate the average improvement in localization with particle filtering. The likelihood function in TAN is realized with Eq.~\mbox{\ref{meth:lklhd}}.} For other approaches in comparison, we use the $softmax$ of cosine similarity to update the particle weights. We experimented on 5 map patches selected with variations in terrain appearance from each dataset.

\begin{table*}
\centering
\begin{tabular}{c|ccccc|ccccc} 
\toprule
                   & \multicolumn{5}{c|}{Scott Reef}                & \multicolumn{5}{c}{Kempten}\\ & Patch 1 & Patch 2  & Patch 3 & Patch 4 & Patch 5   & Patch 1& Patch 2  & Patch 3 & Patch 4& Patch 5\\ 
\midrule
Triplet~\cite{vo2016localizing}& {\color{gray} -5.2\%} & 1.3\% & {\color{gray} -1.8\%} & {\color{gray} -13.3\%}& {\color{gray} -1.1\%} & 5.1\%  & 6.5\% & 12.3\% & 6.6\%& 25.4\%  \\
Triplet (ResNet18) &45.7\% &\textbf{33.2\%} & 18.4\% & 12.1\%& 8.1\%& 31.1\% & 19.13\% & 22.8\%  & 0.1\%& 13.3\% \\
CVM-Net~\cite{cros2018cvmnet} & {\color{gray} -5.6\%} & 21.7\% & 9.3\% &{\color{gray} -0.2\%}& 23.0\%& 30.2\% & 30.9\% & 23.2\%  & 28.15\%& \textbf{44.91\%}  \\
CVM-Net (ResNet18) & 23.9\% & 8.5\% & \textbf{31.5\%} & 7.0\% & 14.6\% & 43.3\% & 15.8\% & 13.1\% & \textbf{29.4}\%& 22.9\% \\
Ours & \textbf{58.4\%} & 18.5\% & 25.7\% & \textbf{25.05\%} & \textbf{25.09\%} & \textbf{46.6\%} & \textbf{37.9\%} & \textbf{28.2\%} & 15.4\%& 19.9\% \\
\bottomrule
\end{tabular}    
\caption{Error reduction on selected map patches. Negative values in grey mean that the error is increased.}
\label{exp:tab_pf}
\end{table*}
\begin{table*}
\centering
\begin{tabular}{ccccc} 
\toprule
                & \begin{tabular}[c]{@{}c@{}}Computation\\Time\end{tabular} & \begin{tabular}[c]{@{}c@{}}GPU Memory\\Consumption\end{tabular} & \begin{tabular}[c]{@{}c@{}}Belief Map  Size\\(Theoretical)\end{tabular} & \begin{tabular}[c]{@{}c@{}}Belief Map Size\\(\hlcblue{in 32-bit float point})\end{tabular}  \\ 
\midrule
Triplet~\cite{vo2016localizing}  & \hlcblue{220.67 s}  & 1.97 Gb  & $\bm{m}\cdot \bm{c}$  &  \textbf{5.01 Mb} \\
Triplet (ResNet18) & \hlcblue{587.19 s}  & 1.61 Gb  & $\bm{m}\cdot \bm{c}$  &  \textbf{5.01 Mb} \\
CVMNet~\cite{cros2018cvmnet}    & \hlcblue{1054.63 s}  & 1.13 Gb  & $m \cdot c\cdot d$  & 2560.06 Mb\\
CVMNet (ResNet18) & \hlcblue{1770.64 s}& 1.70 Gb & $m \cdot c\cdot d$    & 2560.06 Mb\\
Ours & \hlcblue{\textbf{53.81 s}}  & \textbf{1.05 Gb} & $\bm{m}\cdot \bm{c}$ & \textbf{5.01 Mb} \\
\bottomrule
\end{tabular}
\caption{Computational efficiency: $m$, $c$, and $d$ are the size of the belief map, channels and descriptors respectively.}
\label{tab:comp}
\end{table*}

\par We visualize the particle filtering on selected patch 1 from each of both datasets in Fig.~\ref{exp:pf_img}. We see that for both patches, estimation errors along the trajectory are generally reduced with particle filtering, which means that our proposed network learns an effective representation for inference across the scale ratio and modality gap.
However, from the visualization, we also observe certain outlying sequences not converging to ground truth trajectory. The reason behind this can be the inconsistent representation pushing the estimation away from the true trajectory. Also, with no variation in observation nor map, particle filtering is only adding uncertainty into the system. In such cases, the drift is exaggerated, raising the concern about deployment in real environments.
\par The correction performances by particle filtering on all selected map patches are presented in Table~\ref{exp:tab_pf}. For each sequence, we use the accumulated L2 error as the evaluation metric. To eliminate outliers, we report the median accumulated error among all the sequences for each map patch instead of the average error. It can be seen that our approach outperforms other approaches on three trajectories out of five on both datasets. As the particle filter usually has unpredictable performance with different maps or trajectories, we notice that our approach is one of the most stable which reduced average error by $15\%-60\%$ on all selected patches.
\subsubsection{Computational cost and efficiency}
Computation resource consumption of encoding a single map patch is presented in Table~\ref{tab:comp}. The experiment is conducted on a \hlcblue{Jetson Nano computer with 2Gb memory in MAXN power mode.} Our approach is the most efficient regarding the time and memory consumed. It is because other approaches conduct sliding window search which process a large batch of image tiles in parallel, while our approach takes the whole map patch as input. \hlcgreen{Given the limited travelling speed of a mapping robot, our time consumption which is less than 1 minute per map patch promises real-time map encoding.} The belief map size of our approach is the same compact size as that of Triplet network. CVM-Net uses a high dimensional descriptor for deep features of each category, hence the map size is multiplied by descriptor size. The compact map size of our approach enables the potential application with real-time map streaming over the network. 

\section{CONCLUSIONS}
\label{sec:conclusions}
\par In our proposed cross-scale geo-localization task, we break the limitations of existing studies which rely on large-scale high-resolution satellite images as maps.
Instead, we extend the range of modalities that can serve as a map in geo-localization.
We build two datasets from two domains, underwater and aerial, across different modalities and platforms.
On both datasets, our proposed framework demonstrates the ability to learn consistent representations from image observations and maps. Especially, our approach managed to deal with the significant scale ratio between them.
In contrast to previous studies, we move away from the paradigm of image retrieval and exhaustive search.
We encode a map patch into a belief map, which results in the best computation efficiency regarding time, memory, and storage consumption. Such properties make our approach a solution where a map needed to be updated in real-time. We also believe that for small-scale maps, a pixel-wise encoder looking at the whole map is better at extracting and corresponding features across the scales, which is evidenced by the comparison on recall rate. Experiments with synthetic trajectories show that representations learned with our approach are the most effective in localization. Also, the probabilistic nature of our cross-scale representation makes it compatible as a plug-in module in state estimation pipelines.
\par This study overall provides an idea for localizing a perceptual robot in the field with a map. To train the system to describe the map and observations with abstract \say{language}, what we need is just image observations tagged with 2D geospatial locations in the map. No labelling is needed at all as our approach is based on contrastive learning. \hlcgreen{For subsea applications where geo-tags are absent, we have demonstrated how to generate dataset from underwater survey data and train the network in the map's local geo-coordinate.}
\par \hlcgreen{However, a belief map with 5Mb size is still heavy for low-bandwidth networks, e.g. underwater acoustic communication. In the future study, we will be interested in compressing the belief map using its sparsity for efficient map transmitting,} and demonstrate this work on real robots.

\section*{ACKNOWLEDGEMENTS}
\label{sec:acknowledgements}
We gratefully acknowledge the Australian Centre for Field Robotics' Marine Robotics Group for providing the data.



\renewcommand{\bibfont}{\normalfont\small}
\printbibliography


\end{document}